\documentclass[letterpaper, 10pt, conference]{IEEEconf}
\IEEEoverridecommandlockouts
\usepackage{array}
\usepackage{textcomp}
\usepackage{stfloats}
\usepackage{hyperref}
\usepackage{verbatim}
\usepackage{graphicx}
\usepackage{cite}
\usepackage{amsmath,amssymb,amsfonts}
\usepackage{mathtools}
\usepackage{graphicx}
\usepackage{textcomp}
\usepackage[dvipsnames]{xcolor}
\usepackage{balance}
\usepackage[normalem]{ulem}
\usepackage{upgreek}
\usepackage{comment}
\usepackage[formats]{listings}
\usepackage{subcaption}
\usepackage{multirow}
\usepackage{bm}
\usepackage{stmaryrd}
\usepackage{algorithmic}
\usepackage{siunitx}
\usepackage{minted}
\usepackage{csquotes}
\usepackage[color=black,opacity=1,angle=0,scale=1]{background}

\begin{document}

    \title{\LARGE \bf 
        The Constitutional Filter: \\
        Bayesian Estimation of Compliant Agents
    }
    
    \author{
        Simon Kohaut$^{1,*}$, Felix Divo$^{1,*}$, Benedict Flade$^{2}$, \\Devendra Singh Dhami$^{3}$, Julian Eggert$^{2}$,  Kristian Kersting$^{1, 4, 5, 6}$
        \thanks{$^{*}$ Authors contributed equally}
        \thanks{
            $^{1}$ Artificial Intelligence and Machine Learning Group, \newline\hspace*{1.6em} 
            Department of Computer Science, \newline\hspace*{1.6em}
            TU Darmstadt, 64283 Darmstadt, Germany 
        }%
        \thanks{
            $^{2}$ Honda Research Institute Europe GmbH, \newline\hspace*{1.6em} 
            Carl-Legien-Str. 30, 63073 Offenbach, Germany
        }%
        \thanks{
            $^{3}$
            Uncertainty in Artificial Intelligence Group, \newline\hspace*{1.6em}
            Department of Mathematics and Computer Science, \newline\hspace*{1.6em}
            TU Eindhoven, 5600 MB Eindhoven, Netherlands%
        }%
        \thanks{
            $^{4}$ Hessian AI
        }%
        \thanks{
            $^{5}$ Centre for Cognitive Science
        }%
        \thanks{
            $^{6}$ German Center for Artificial Intelligence (DFKI)
        }%
    }

    \backgroundsetup{
      contents={
          \begin{tikzpicture}
            \node at (current page.center) [align=center] {\textcopyright 2025 IEEE. Personal use of this material is permitted. \\ Permission from IEEE must be obtained for all other uses, in any current or future media, including reprinting/republishing this material for advertising or promotional purposes, \\ creating new collective works, for resale or redistribution to servers or lists, or reuse of any copyrighted component of this work in other works.\\
            DOI: 10.1109/IROS60139.2025.11246215};
        \end{tikzpicture}},
      placement=bottom,
      scale=0.6,
      vshift=20
    }
    
    \maketitle
    \thispagestyle{empty}
    \pagestyle{empty}
    
    \begin{abstract}
    Predicting agents impacted by legal policies, physical limitations, and operational preferences is inherently difficult.
    In recent years, neuro-symbolic methods have emerged, integrating machine learning and symbolic reasoning models into end-to-end learnable systems.
    Hereby, a promising avenue for expressing high-level constraints over multi-modal input data in robotics has opened up.
    This work introduces an approach for Bayesian estimation of agents expected to comply with a human-interpretable neuro-symbolic model we call its Constitution.
    Hence, we present the Constitutional Filter (CoFi), leading to improved tracking of agents by leveraging expert knowledge, incorporating deep learning architectures, and accounting for environmental uncertainties.
    CoFi extends the general, recursive Bayesian estimation setting, ensuring compatibility with a vast landscape of established techniques such as Particle Filters.
    To underpin the advantages of CoFi, we evaluate its performance on real-world marine traffic data. 
    Beyond improved performance, we show how CoFi can learn to trust and adapt to the level of compliance of an agent, recovering baseline performance even if the assumed Constitution clashes with reality.
\end{abstract}


    \section{Introduction}
\label{sec:introduction}

\PARstart{P}{redicting} a moving target without knowing its intentions or control methodology is inherently challenging.
Specifically, when limited to the most basic state-space models, e.g., constant velocity or acceleration systems, only a vague understanding of the agent's trajectory can be formed.
When this is the case, e.g., with human-operated vehicles such as cars or boats, one can leverage knowledge about the agent's environment to restrict the predicted state evolution.
For instance, in the case of road vehicles, one can inform the forward model with the curvature of the road the agent is currently traversing~\cite{Toledo2010}.
However, the insight into purely geometric aspects does not fully leverage the frequently available background knowledge on legislative and logical constraints.

\begin{figure}[t]
    \centering
    \includegraphics[width=\linewidth]{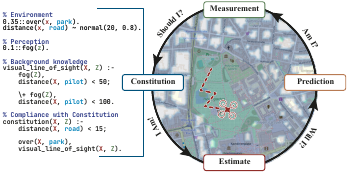}
    \caption{
        \textbf{The Constitutional Filter's estimation cycle:}
        A Constitution for a remote-controlled, unmanned aircraft system is shown on the left.
        The right-hand side shows the tracked drone and the likely compliant areas shaded in blue.
    }
    \label{fig:motivation}
\end{figure}

To this end, we present the Constitutional Filter (CoFi), a novel prediction model coupling an agent's motion to a neuro-symbolic model of its internal rules and perception of its state space.
While such models have been employed as the basis for planning, e.g., mission design for Unmanned Aircraft Systems in regulated urban environments~\cite{kohaut2023mission}, this work explores how to apply such techniques for predicting other agents' motion without insights into their actual control and intentions.
More specifically, we employ probabilistic first-order logic, capturing knowledge about the legal and physical constraints in an uncertainty-aware environment representation.
This permits expressing rules over spatial relations between the agent's state-space and geographic features from potentially inaccurate map data.

In summary, our key contributions are:
\begin{itemize}
    \item We present the Constitutional Filter (CoFi), an extension to Bayesian estimation that tracks compliant agents informed by a neuro-symbolic model of the environment's and agent's rules as illustrated in Figure~\ref{fig:motivation}.
    \item We demonstrate how CoFi learns to trust observed agents, quantifying the degree to which they are expected to act according to the assumed Constitution.
    \item We show how CoFi adapts the impact of the Constitution according to the learned trust, assuring that a faulty constitution does not degrade filter performance.
\end{itemize}
Additionally, we provide an open-source implementation of CoFi as part of our framework for Probabilistic Mission Design in multi-modal mobility\footnote{\href{https://www.github.com/HRI-EU/ProMis}{github.com/HRI-EU/ProMis}}.

    \section{Related Work}
\label{sec:related}

CoFi is a novel approach for the Bayesian estimation of compliant agents.
To contextualize CoFi, we discuss neuro-symbolic systems (reasoning), probabilistic robotics (filtering), as well as perception and representation (mapping).

\subsection{Neuro-Symbolic Systems}

Neuro-symbolic systems are an emerging field aiming to intertwine programmatic reasoning on a symbolic level with the sub-symbolic capabilities of deep learning models.

One of the earliest programmatic reasoning systems based on first-order logic is Prolog~\cite{colmerauer1990introduction}.
To additionally embrace uncertainty into programmatic logic, systems such as Bayesian Logic Programs~\cite{bayesian_logic} and Probabilistic Logic Programs~\cite{problog,inference_in_plp} have been introduced.
While they were not formulated for end-to-end learning with artificial neural networks, languages such as DeepProbLog~\cite{deepproblog}, NeurASP~\cite{neurasp}, and SLASH~\cite{slash} close this gap and combine the strengths of neural information processing and probabilistic reasoning.
Recent work has demonstrated how such systems can be employed as a basis for Probabilistic Mission Design~\cite{kohaut2023mission,kohaut2024ceo} by encoding, e.g., traffic laws in hybrid probabilistic first-order logic~\cite{nitti2016probabilistic} as a basis for planning or granting clearance to an agent.

CoFi follows the idea of encoding the laws and regulations that govern an agent's environment in a neuro-symbolic setting, formulating the agent's Constitution.
In contrast to ProMis, this enables CoFi to integrate state estimation, sensor data, and high-level semantic information into a Bayes Filter for tracking targets with high-level rules of operation.

\subsection{Probabilistic Robotics}

One cannot deny the critical role of probabilistic methods in robotics, commonly based on Bayesian beliefs~\cite{Thrun2005}.
For instance, state-of-the-art localization usually relies on employing members of the family of Bayes Filters (Section~\ref{sec:bayesian_filters}), such as the Kalman Filter~\cite{kalman1960new}, providing a closed-form solution to recursive tracking in linear and Gaussian systems.

To enable tracking in non-linear systems while keeping the Gaussian assumption, generalizations such as the Extended Kalman Filter~\cite{Davison2007,Junior2022,Fang2022} and the Unscented Kalman Filter~\cite{Martinez2005,Kuti2023} have been introduced.
While the former relies on a linearization of the model, the latter works by forwarding samples through the so-called unscented transform to conserve the Gaussian nature of the estimate. 
When dropping linearity and distribution assumptions on the underlying process, one can employ the Particle Filter~\cite{del1997nonlinear,Gao2020} instead.
Here, approximate results are obtained using sequential Monte Carlo techniques.

CoFi is also a member of the Bayes Filters.
By introducing a third step to the paradigm of prediction-correction schemes, namely the Constitution step, CoFi guides the estimation process through the likelihood of a satisfied Constitution given the state, measurement, and high-level semantic information.
Because CoFi does not alter the roles of the process and measurement models, it stays compatible with a wide range of prior work on filter techniques that can complement CoFi's insights based on probabilistic first-order logic. 

\subsection{Perception and Environment Representation}

Numerous sensors have been developed to perceive oneself and the environment.
They range from proprioceptive sensors, e.g., the Global Navigation Satellite System~\cite{joubert2020developments} and Inertial Measurement Units~\cite{Zhang2012,Gao2022} which inform about the agent themselves, to exteroceptive sensors such as cameras~\cite{Alkendi2021,Flade2018}, Lidar~\cite{Demir2019,Petrlik2021}, and Radar~\cite{Ward2016,Michalczyk2022} which inform about the environment.

These perceptions can be leveraged to improve filter performance significantly~\cite{Karaked2022}.
For instance, in the autonomous driving context, this has been achieved by coupling observations of lane markings~\cite{Flade2020} or stop lines~\cite{Barth2013} with a vehicle's proprioception. 
Similarly, knowledge of the road geometry can be employed to successfully predict vehicle motion along a local Frenet coordinate system~\cite{Toledo2010}, assuming the agent follows local traffic regulations.
Likewise, high-level semantic information on the road infrastructure and rules can be leveraged to improve motion prediction~\cite{Svensson2016}, e.g., lane count and directionality.
Like transportation systems need to follow road laws, tracking pedestrians profits from integrating data about local laws and environment features such as movement patterns and social interactions~\cite{Krishanth2017}.

Along these lines of research, CoFi integrates high-level concepts and semantics of the environment by employing a Statistical Relational Map~\cite{flade2023star} (StaR Map, see Section~\ref{sec:starmaps}), assigning heterogeneous levels of uncertainty to geographic features~\cite{flade2021error}.
Hence, in its Constitution, probabilistic inference across likely states and measurements results in an additional distribution weighting the estimation process, encoding how likely the agent's local laws and assumed preferences are satisfied.

    \section{Preliminaries}
\label{sec:problem}

We present the general formulations and notation of recursive estimation of dynamic systems in a Bayesian setting and the uncertainty-aware, semantic environment representation of Statistical Relational Maps, as they will be required later in Sec.~\ref{sec:method} for building the Constitutional Filter.

\subsection{Bayesian Filters}
\label{sec:bayesian_filters}
Bayesian filters recursively estimate the state $\mathbf{x}_t \in \mathcal{X}$ of a dynamical system at time $t \in \mathbb{N}$. 
While the state is hidden, a combination of prediction through a process model $f$ with control inputs $\mathbf{u}_t \in \mathbb{R}^U, U \in \mathbb{N}$ and belief updates through an observation model $h$ with measurements $\mathbf{z}_t \in \mathcal{Z}$ is employed to keep track of $\mathbf{x}_t$.
More specifically, the system can be written as 
\begin{align}%
    \mathbf{x}_t = f(\mathbf{x}_{t-1}, \mathbf{u}_t, \mathbf{e}_{x,t}) \text{ and }
    \mathbf{z}_t = h(\mathbf{x}_t, \mathbf{u}_t, \mathbf{e}_{z,t}).
\end{align}%

Intuitively speaking, $f$ expresses an expectation on a state's evolution within a discrete time-step, while $h$ encodes the values produced by a sensor.
Here, both models are subject to i.i.d. noise $\mathbf{e}_{x,t}$ and $\mathbf{e}_{z,t}$ respectively.
Depending on the application, this formulation can slightly change, e.g., often the input $\mathbf{u}_t$ is not part of the computation of the system's output $\mathbf{z}_t$.
We will omit $\mathbf{u}_t$ for simplicity in the following. 

A Bayesian filter operates in two steps.
First, a prior belief is obtained using the last estimate $\mathbf{x}_{t-1}$ as
\begin{align}
    p(\mathbf{x}_t | \mathbf{z}_{1:t-1}) = \int \overbrace{p(\mathbf{x}_t | \mathbf{x}_{t-1})}^{\mathclap{\text{Process Model } f}} p(\mathbf{x}_{t-1} | \mathbf{z}_{1:t-1}) d\mathbf{x}_{t-1}.
    \label{eq:prior}
\end{align}
Second, once a measurement $\mathbf{z}_t$ is available, the prediction is updated using Bayes' rule
\begin{align}
    p(\mathbf{x}_t | \mathbf{z}_{1:t}) &= \frac{\overbrace{p(\mathbf{z}_t | \mathbf{x}_t)}^{\mathclap{\text{Observation Model } h}} p(\mathbf{x}_t | \mathbf{z}_{1:t-1})}{p(\mathbf{z}_t | \mathbf{z}_{1:t-1})}\text{, where} 
    \label{eq:bayes_update}
    \\
    p(\mathbf{z}_t | \mathbf{z}_{1:t-1}) &= \int p(\mathbf{z}_t | \mathbf{x}_{t}) p(\mathbf{x}_{t} | \mathbf{z}_{1:t-1}) d\mathbf{x}_t.
    \label{eq:normalization}
\end{align}%
As is the case for the Kalman filter, closed-form solutions to the equations above exist depending on assumptions about the observed process, e.g., linear state transitions and Gaussian noise.
Without such strong assumptions, approaches such as the Particle filter allow the tracking of the hidden state using sampling techniques.

As a final remark, in recursive filtering, we assume the Markov property for states and measurements, i.e.,
\begin{align}
p(\mathbf{x}_t | \mathbf{x}_{1:t-1}, \mathbf{z}_{1:t-1}) &= p(\mathbf{x}_t | \mathbf{x}_{t-1}) \text{ and } \\
p(\mathbf{z}_t | \mathbf{x}_{1:t}) &= p(\mathbf{z}_t | \mathbf{x}_t).
\end{align}
Intuitively, the previous state contains sufficient information for prediction to discard all other historical data.

\subsection{Statistical Relational Maps}
\label{sec:starmaps}

Statistical Relational Maps (StaR Maps) have been recently introduced to represent uncertain environments consisting of semantic features in a relational and probabilistic manner~\cite{flade2023star}.
Rather than containing a graphical representation of the environment, a StaR Map parameterizes hybrid probabilistic (discrete and continuous) spatial relations.

\begin{figure}
    \centering
    \begin{subfigure}{0.525\linewidth}
        \includegraphics[width=\textwidth]{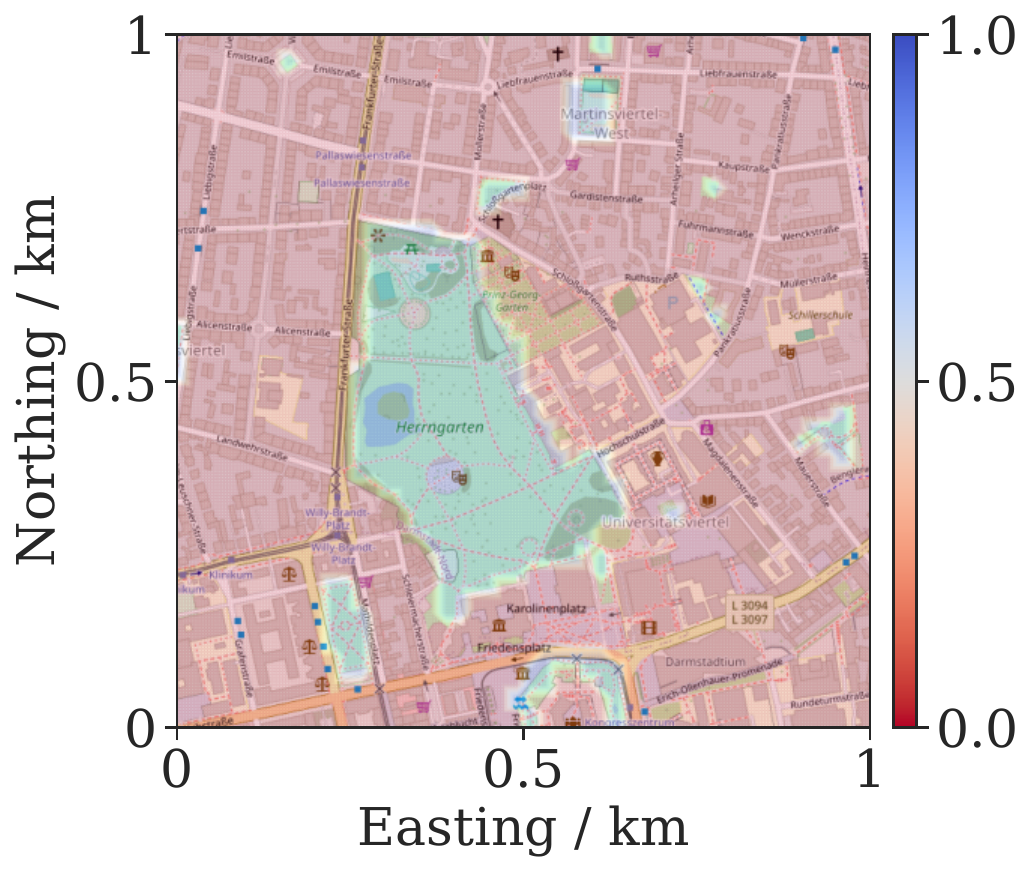}
        \caption{$E\left[over(X, park)\right]$}
    \end{subfigure} 
    \hfill
    \begin{subfigure}{0.445\linewidth}
        \includegraphics[width=\textwidth]{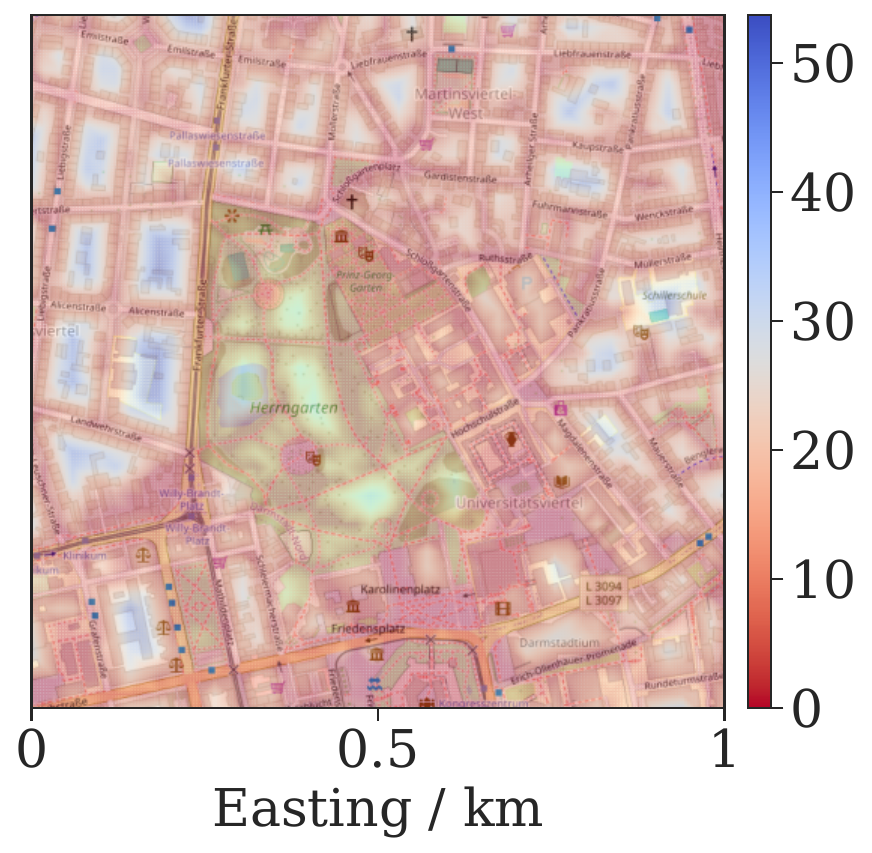}
        \caption{$E\left[distance(X, road)\right]$}
    \end{subfigure}
    \caption{
        \textbf{StaR Maps parameters in an Advanced Aerial Mobility setting}:
        Here, the expected values of two basic probabilistic spatial relations are shown in an urban environment, namely \textit{over} and \textit{distance} 
        (used in Figure~\ref{fig:motivation}).
    }
    \label{fig:spatial_relations}
\end{figure}

Using a StaR Map, the so-obtained spatial relations can be translated to distributional ground atoms in first-order logic, exemplified by the Listing in Figure~\ref{fig:motivation}.
Furthermore, one can visualize the probabilistic spatial relations, e.g., by computing the expected values across the mapped space as shown in Figure~\ref{fig:spatial_relations}.

Consider the exemplified relations \textit{over(x, g)} and \textit{distance(x, g)} between a point $\mathbf{x} \in \mathbb{R}^D$ to a set of environment features $g \in \mathcal{G}$.
The specifics, such as the dimensionality of the mapped space and the sets of features, will depend on the application. 
For example, in the aerial mobility scenario illustrated in Figure~\ref{fig:motivation}, $\mathcal{G} = \{park, road, pilot\}$ and $D=2$.
Let us discuss how StaR Maps estimate the parameters of such relations.

Consider a map $\mathcal{M} = (\mathcal{V}, \mathcal{E}, \rho)$ be a triple of vertices $\mathcal{V}$, edges $\mathcal{E}$, and a tagging function $\rho : \mathcal{V} \to \mathcal{P}(\mathcal{G})$.
The function $\rho(\mathbf{v}) \subseteq \mathcal{G}$ annotates vertex $\mathbf{v} \in \mathcal{V}$ with a set of semantic tags.
If a path exists between two vertices in $\mathcal{V}$ across edges in $\mathcal{E}$, StaR Maps consider them part of the same \textit{feature}, hence $\rho$ assigns the same type to each.

To address inaccuracies of the map data, e.g., due to low-cost sensors or crowd-sourced data, StaR Maps employ a stochastic error model analogously to prior work~\cite{flade2021error}.
That is, for each $\mathbf{v}_{i, j} \in \mathcal{V}$, being the $j$-th vertex of the $i$-th feature, the following generates $N \in \mathbb{N}$ samples:
\begin{align*}
    \mathbf{\Phi}^{(n)} &\sim \phi_i \tag*{(Transformation)} \\
    \mathbf{t}^{(n)} &\sim \kappa_i \tag*{(Translation)} \\
    \mathbf{v}^{(n)}_{i,j} &= \mathbf{\Phi}^{(n)} \cdot \mathbf{v}_{i, j} + \mathbf{t}^{(n)} \tag*{(Generation)}
\end{align*}
Here, $\phi_i$ and $\kappa_i$ are feature-wise distributions of linear maps $\mathbf{\Phi}^{(n)}$, e.g., rotation, scaling or shearing, and translations $\mathbf{t}^{(n)}$.
With $\mathcal{M}^{(n)}$ being generated by taking the $n$-th sample of each vertex and a copy of the edges $\mathcal{E}$, this collection of randomized maps allows StaR Maps to compute the statistical moments of its spatial relations.

Let $r(\mathcal{M}^{(n)}, \mathbf{x}, g)$ be a deterministic function evaluating a spatial relation on map $\mathcal{M}^{(n)}$ at location $\mathbf{x} \in \mathbb{R}^d$ limited to features for which $\rho(\mathbf{v}) = g$.
Through this sampling process, we can empirically estimate statistical moments, e.g., mean and variance, with respect to the chosen relation $r$, location $\mathbf{x}$, and tag $g$:%
\begin{align}
    \widehat{\mu_r} &= \frac{1}{N} \sum_n r(\mathcal{M}^{(n)}, \mathbf{x}, g) \\
    \widehat{\sigma^2_r} &= \frac{1}{N - 1} \sum_n \left(r(\mathcal{M}^{(n)}, \mathbf{x}, g) - \widehat{\mu_r}\right)^2
\end{align}%
For example, assume \textit{distance}$(\mathbf{x}, g) \sim \mathcal{N}(\mu_d, \sigma^2_d)$ to be a normally distributed random variable with the deterministic function $r_d$.
That is, the function $r_d(\mathcal{M}^{(n)}, \mathbf{x}, g)$ computes the Euclidean distance from $\mathbf{x}$ to the closest feature in $\mathcal{M}^{(n)}$.

\section{Methods}
\label{sec:method}

We present the Constitutional Filter (CoFi) for tracking compliant agents with unknown internal control and planning mechanisms.
This includes the discussion of (\hyperref[sec:method:arch]{\textit{A}}) CoFi's novel architecture, introducing neuro-symbolic concepts to recursive Bayesian estimation, (\hyperref[sec:method:constitution]{\textit{B}}) the construction and (\hyperref[sec:method:inference]{\textit{C}}) semantics of the Constitution as a probabilistic reasoning model, and (\hyperref[sec:method:density]{\textit{D}}) how to obtain a density of the constitutional likelihood to be integrated into (\hyperref[sec:method:update]{\textit{E}}) the constitutional belief update.
Finally, we show how (\hyperref[sec:method:trust]{\textit{F}}) CoFi learns and integrates a level of trust about how compliant observed agents are.

\subsection{Architecture}
\label{sec:method:arch}

\begin{figure}[t]
    \centering
    \includegraphics[width=\linewidth]{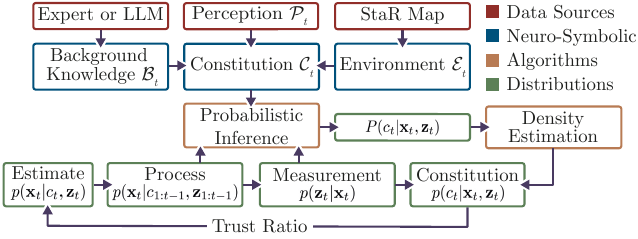}
    \caption{
        \textbf{The Constitutional Filter's architecture:}
        CoFi extends recursive Bayesian estimation using a neuro-symbolic model of the tracked agent's decision-making constraints, integrating background knowledge, perception, and a probabilistic environment representation.
    }
    \label{fig:architecture}
\end{figure}

As illustrated in Figure~\ref{fig:architecture}, CoFi integrates neuro-symbolic representations of the agent's internal preferences, physical limitations, and environment policies, such as traffic laws, into recursive Bayesian estimation.
Hence, CoFi leverages high-level expert knowledge about the task, environment, and semantic information beyond simple sensor readings.
Specifically, besides a process model $f$ and observation model $h$, CoFi is equipped with the Constitution $\mathcal{C}_t$, allowing it to map states and measurements into the probability of the Constitution being satisfied.
Through a learned trust ratio $\tau$, CoFi automatically regulates the influence the Constitution has on the belief update, recovering the equations of an unconstitutional filter if the observed agent proves incompliant.

\subsection{Constitutions as Deep Probabilistic Logic Programs}
\label{sec:method:constitution}

We encode the Constitution as a deep probabilistic first-order logic program $\mathcal{C}_t = \mathcal{B}_t \cup \mathcal{P}_t \cup \mathcal{E}_t$, combined of sub-programs encoding background knowledge $\mathcal{B}_t$, perception $\mathcal{P}_t$, and environment representation $\mathcal{E}_t$ as illustrated in Figure~\ref{fig:motivation}.
For CoFi, we assume $\mathcal{B}_t$ to be provided by a domain expert, i.e., by formalizing their knowledge as first-order logic or by translating natural language descriptions through, e.g., a Large Language Model.
Further, $\mathcal{P}_t$ encodes the semantics of the environment as experienced during the agent's lifetime, e.g., information from other agents in the environment.
Finally, $\mathcal{E}_t$ comprises spatial relations queried from a StaR Map as outlined in Section~\ref{sec:starmaps}.

Constitutions consist of \textit{clauses}, each being made up of a \textit{head}, optional \textit{body}, and \textit{distribution}.
Consider the following two clauses.
\begin{align}
    \label{eq:clauses}
    p \ :: \ &r_1(a_1, \ldots, a_n) \ \text{:-} \ l_1,\ \ldots,\ l_m. \tag*{(Categorical)} \\
    &r_2(a_1, \dots, a_i) \sim p(\mathbf{\theta}) \ \text{:-} \ l_1,\ \ldots,\ l_j. \tag*{(Continuous)}
    \label{eq:distributional_clauses}
\end{align}
In the first case, the head $r_1$ is true with a probability $p$ given that all the \textit{literals} $l_k$ in the body are true.
Analogously, in the second case, head $r_2$ is distributed according to the density $p(\mathbf{\theta})$ with parameters $\mathbf{\theta}$ if its body is true.
If the right-hand side is empty, the head is regarded as a fact and distributed independently of any other symbols.

For example, let us consider the running examples from Section~\ref{sec:starmaps} as probabilistic clauses:
\begin{align}
    \label{eq:example_clause_over}
    &0.95 \ :: \ \text{over}(\text{x}, \text{park}). \\
    &\text{distance}(\text{x}, \text{road}) \sim normal(100, 1).
    \label{eq:example_clause_distance}
\end{align}
This means that at the location referenced with the term x, there is a park with a probability of $0.95$, and the distance to the nearest road is expected to be about $\SI{100}{\meter}$.

A logic program must be solved, i.e., the models under which a chosen clause is satisfied must be found to perform probabilistic inference.
While details will differ across off-the-shelf solving pipelines such as Prolog~\cite{prolog} or clingo~\cite{gebser2017clingo}, the rough process can be described in two steps.
First, the program is grounded, replacing all variables with the possible values from the program's domain.
Second, the solver processes the grounded program, enumerating all solutions consisting of models $j \in \mathcal{J}$ of ground atoms $a \in \mathcal{A}$.

\subsection{Exact Probabilistic Inference}
\label{sec:method:inference}

We now aim to compute the probability of a pair $(\mathbf{x}_t, \mathbf{z}_t)$ satisfying the first-order logic program of $\mathcal{C}_t$ when solving for the \textit{constitution(X, Y)} clause.
For exact probabilistic inference, one needs to assign the probabilities $P(a = j(a) | \mathbf{x}_t, \mathbf{z}_t, \mathcal{C}_t)$ of atom $a$ to take on the value assigned by model $j$ given the current state $\mathbf{x}_t$, measurement $\mathbf{z}_t$ and constitution $\mathcal{C}_t$.
For CoFi, we refer to a StaR Map (see Section~\ref{sec:starmaps}) to provide the distribution parameters to ground atoms expressing spatial relations as in Eqs.~\ref{eq:example_clause_over} and \ref{eq:example_clause_distance}.
Further, we assume $\mathcal{B}_t$ to be parameterized according to expert knowledge or learned and $\mathcal{P}_t$ to encode uncertainties associated with the respective perceived properties (see Figure~\ref{fig:motivation} and Listing~\ref{listing:marine_model}).
In turn, the probability $P(C_t | \mathbf{x}_t, \mathbf{z}_t)$ of the Constitution being satisfied given the state and measurement at time $t$ is then obtained via the sum-product
\begin{align}
    \hspace{-0.2cm}P(C_t | \mathbf{x}_t, \mathbf{z}_t) = \sum\nolimits_{j \in \mathcal{J}} \prod\nolimits_{a \in \mathcal{A}} P(a = j(a) | \mathbf{x}_t, \mathbf{z}_t, \mathcal{C}_t).
    \label{eq:wmc}
\end{align}
A knowledge compiler is often employed to compress this sum-product using a heuristic search for a minimal formula, often leading to substantial inference speedups~\cite{muise2012d}.

\subsection{Density Estimation of the Constitutional Likelihood}
\label{sec:method:density}

Note that $P(C_t | \mathbf{x}_t, \mathbf{z}_t)$ of Eq.~\ref{eq:wmc} is a discrete distribution, encoding the probability of the constitution being satisfied given a state and measurement.
To integrate the Constitution into a Bayesian belief update in CoFi's process, we must find the probability density function $p(c_t | \mathbf{x}_t, \mathbf{z}_t)$ of a sample satisfying the Constitution.
To do so, we compute the set of probabilities of the constitution being satisfied 
\begin{align}
    \mathcal{S} = \big\{ P(C_t | \mathbf{x}_t^{(n)}, \mathbf{z}_t^{(n)})\big\}_{n \in \{ 1, \dots, N \}},
\end{align}
where $\mathbf{x}_t^{(n)} \sim p(\mathbf{x}_t | c_{1:t-1}, \mathbf{z}_{1:t-1})$ and $\mathbf{z}_t^{(n)} \sim p(\mathbf{z}_t | \mathbf{x}_t)$.
One can then approximate the density $p(c_t | \mathbf{x}_t, \mathbf{z}_t)$ from $\mathcal{S}$, for instance, using Kernel Density Estimation~\cite{davis2011remarks,parzen1962estimation}.

\begin{figure*}
    \centering
    \begin{subfigure}{0.2\linewidth}
        \includegraphics[width=\textwidth]{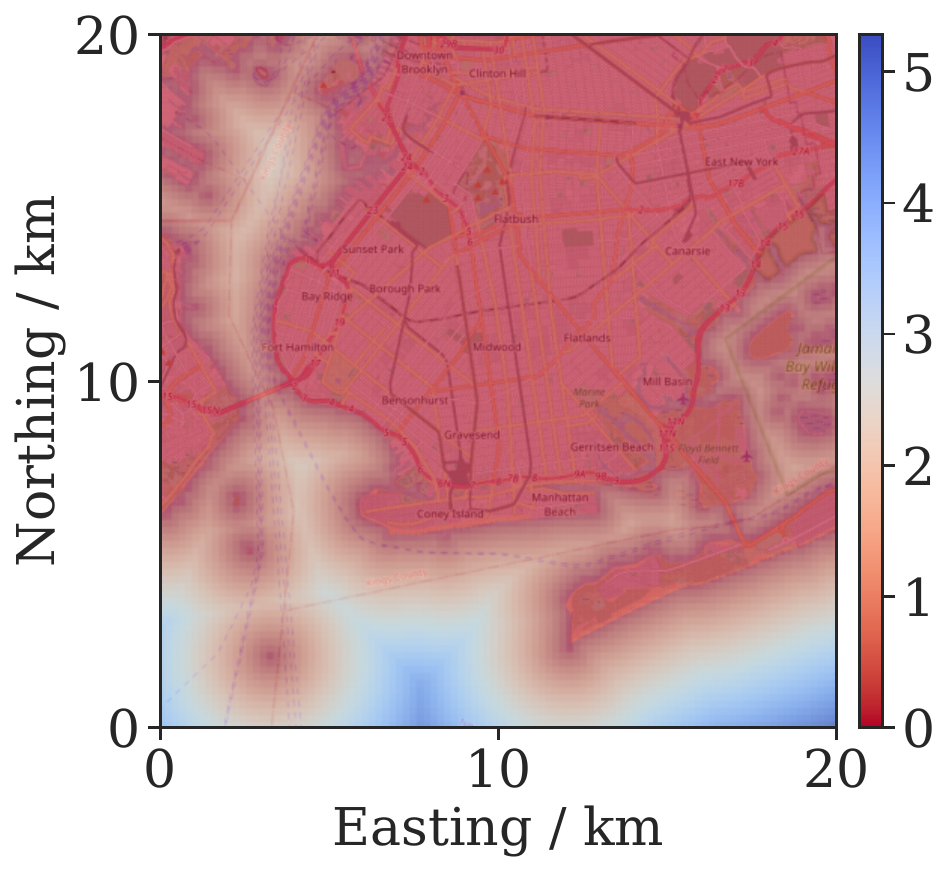}
        \caption{$distance(X, land)$}
    \end{subfigure} 
    \hfill
    \begin{subfigure}{0.19\linewidth}
        \includegraphics[width=\textwidth]{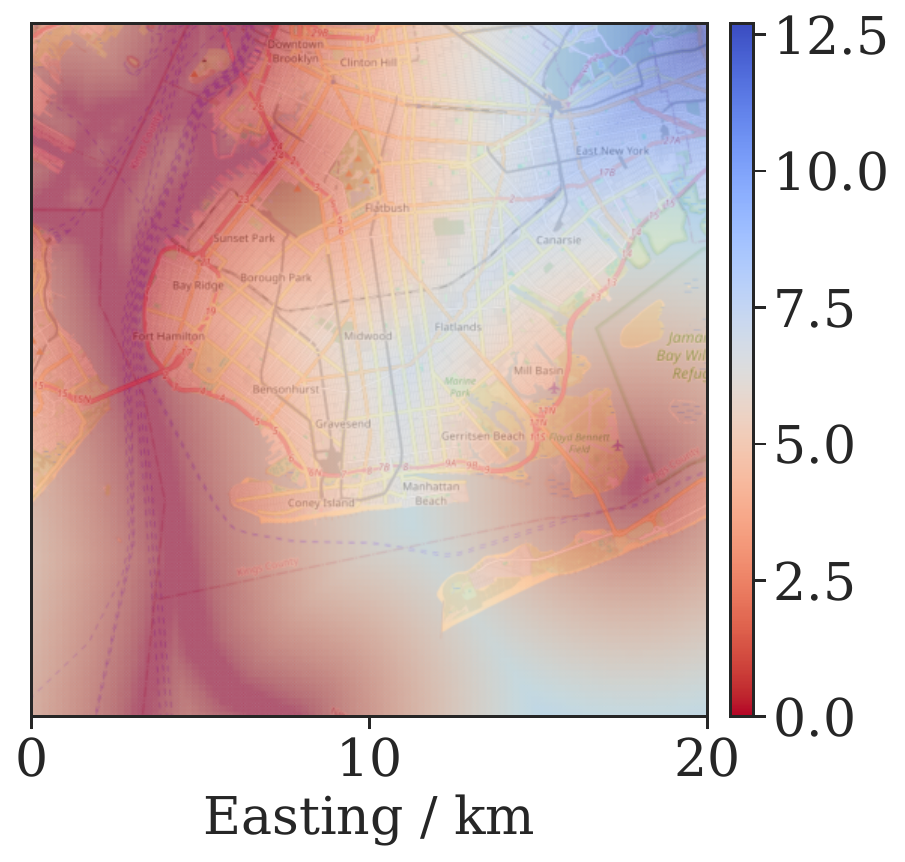}
        \caption{$distance(X, way)$}
    \end{subfigure}
    \hfill
    \begin{subfigure}{0.185\linewidth}
        \includegraphics[width=\textwidth]{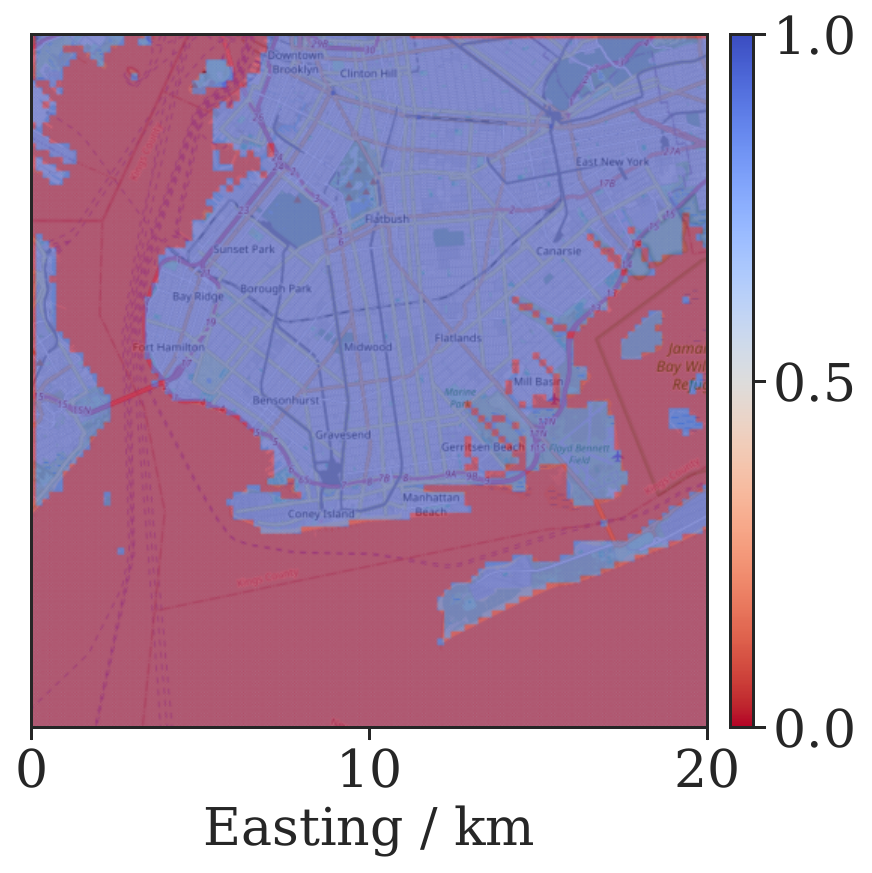}
        \caption{$over(X, land)$}
    \end{subfigure} 
    \hfill
    \begin{subfigure}{0.185\linewidth}
        \includegraphics[width=\textwidth]{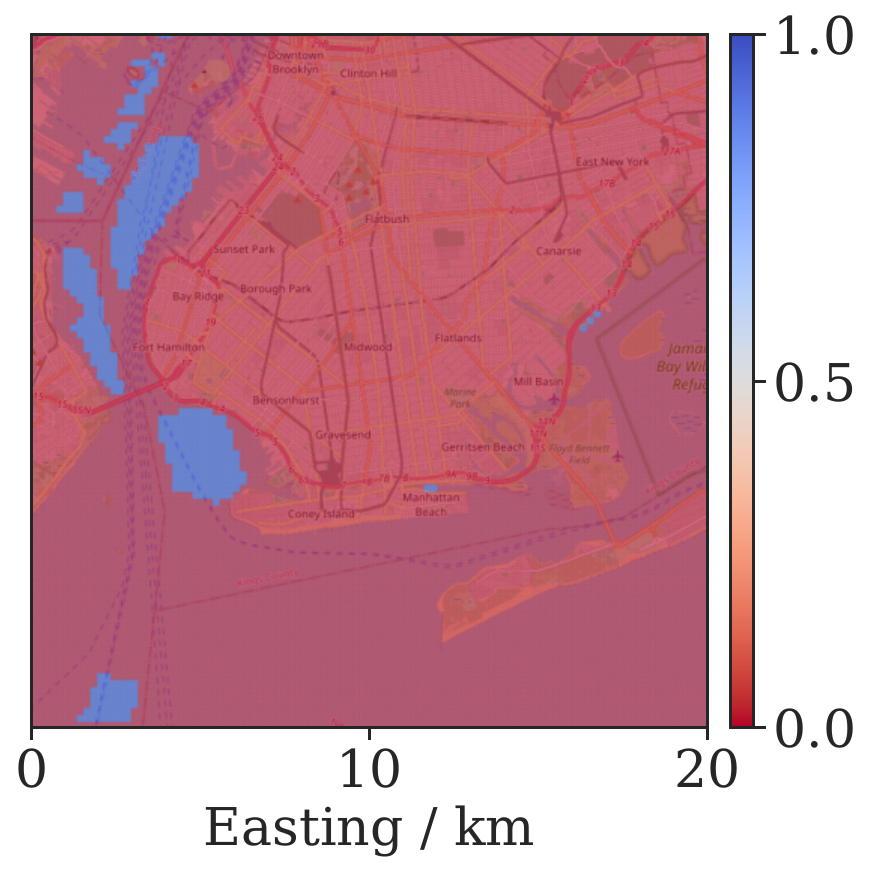}
        \caption{$over(X, anchorage)$}
    \end{subfigure} 
    \hfill
    \begin{subfigure}{0.19\linewidth}
        \includegraphics[width=\textwidth]{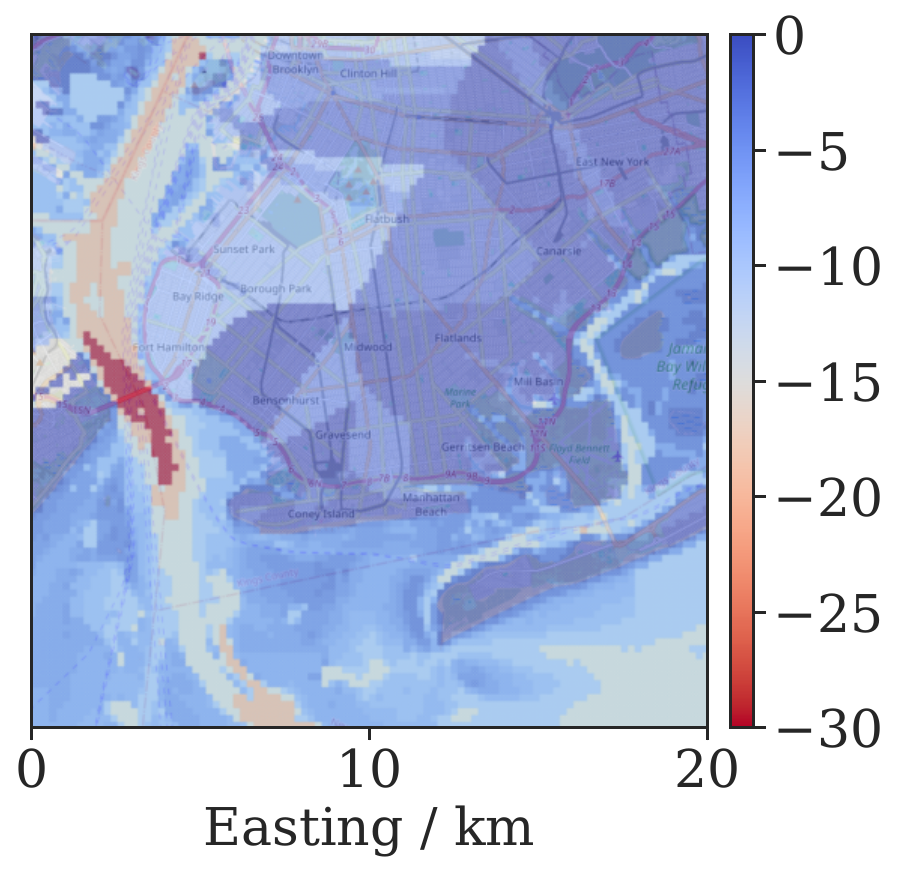}
        \caption{$depth(X, water)$}
    \end{subfigure}
    \caption{
        \textbf{Statistical relational maps for marine traffic:}
        Expectations of StaR Map relations employed in Listing~\ref{listing:marine_model} at New York's harbor, considering land, waterways, anchorage areas, and depth.
        Distances are shown in kilometers.
    }
    \label{fig:star_maps_parameters}
\end{figure*}

\subsection{Constitutional Bayesian Belief Update}
\label{sec:method:update}

We have shown how one can compute the probability that the Constitution is satisfied for a state $\mathbf{x}_t$ and measurement $\mathbf{z}_t$, yielding $p(c_t | \mathbf{x}_t, \mathbf{z}_t)$.
Hence, we can integrate the Constitution into a general Bayesian belief update.

CoFi's prior of the state follows the standard formulation while considering past evaluations of the agent's Constitution, similar to Eq.~\ref{eq:prior}:
\begin{align}
\begin{split}
    \hspace{-0.2cm} p(\mathbf{x}_t | c_{1:t-1}, \mathbf{z}_{1:t-1}) = \int &p(\mathbf{x}_t | \mathbf{x}_{t-1}) \\ &p(\mathbf{x}_{t-1} | c_{1:t-1}, \mathbf{z}_{1:t-1}) d\mathbf{x}_{t-1}.
    \label{eq:cofi_prior}
\end{split}
\end{align}
Further, we obtain the posterior of an agent's location given the Constitution by plugging Eq.~\ref{eq:cofi_prior} into Bayes' rule as in Eq.~\ref{eq:bayes_update}:
\begin{equation}
    p(\mathbf{x}_t | c_{1:t}, \mathbf{z}_{1:t}) = \frac{p(c_t, \mathbf{z}_t | \mathbf{x}_t) p(\mathbf{x}_t | c_{1:t-1}, \mathbf{z}_{1:t-1})}{\int p(c_t, \mathbf{z}_t | \mathbf{x}_t) p(\mathbf{x}_{t} | c_{1:t-1}, \mathbf{z}_{1:t-1}) d\mathbf{x}_t}.
\end{equation}%
Finally, by applying the chain rule of probability to the conditional joint probability $p(c_t, \mathbf{z}_t | \mathbf{x}_t)$, we get the Constitutional Bayesian belief update such that
\begin{align}
\begin{split}
    \hspace{-0.1cm} %
    p(\mathbf{x}_t | c_{1:t}, \mathbf{z}_{1:t}) \hspace{-0.1cm} = \hspace{-0.1cm} \frac{\overbrace{p(c_t | \mathbf{x}_t, \mathbf{z}_t)}^{\mathclap{\text{Constitution}}}\overbrace{p(\mathbf{z}_t | \mathbf{x}_t)}^{\mathclap{\text{Measurement}}} \overbrace{p(\mathbf{x}_t | c_{1:t-1}, \mathbf{z}_{1:t-1})}^{\mathclap{\text{Prediction}}}}{\int p(c_t | \mathbf{x}_t, \mathbf{z}_t) p(\mathbf{z}_t | \mathbf{x}_t) p(\mathbf{x}_{t} | c_{1:t-1}, \mathbf{z}_{1:t-1}) d\mathbf{x}_t}.
    \label{eq:cofi_update}
\end{split}%
\end{align}%
 
\subsection{Constitutional Trust Ratio}
\label{sec:method:trust}

One cannot blindly assume that the proclaimed Constitution perfectly matches an agent's behavior.
First, the Constitution might be incomplete or erroneous.
Second, the agent might not diligently follow the rules of the environment, e.g., by disregarding local policies.
Hence, we introduce the trust ratio $\tau(\bm{\psi})$, allowing CoFi to blend between the Constitutional likelihood and a uniform distribution:
\begin{align}
    \hspace{-0.25cm}p_\tau(c_t | \mathbf{x}_t, \mathbf{z}_t) &= \tau(\bm{\psi}_t) p(c_t | \mathbf{x}_t, \mathbf{z}_t) \hspace{-1pt}+\hspace{-1pt} (1 \hspace{-1pt}-\hspace{-1pt} \tau(\bm{\psi}_t)) U(0, 1) \\
    \bm{\psi}_t &= \left[ \begin{array}{ccc} \psi_1(\mathbf{x}_t, \mathbf{z}_t) & \cdots &\psi_K(\mathbf{x}_t, \mathbf{z}_t) \end{array} \right]
\end{align}
Here, the vector $\bm{\psi}$ captures trust features, e.g., the type of a vehicle or its current velocity, that the trust depends on.

The density $p_\tau(c_t | \mathbf{x}_t, \mathbf{z}_t)$ can then replace the standard Constitutional likelihood in Eq.~\ref{eq:cofi_update}.
Note that for $\tau=1$, we fully trust the correctness of the constitution, and for $\tau=0$, we obtain a standard Bayesian belief update that discards the constitution entirely (unconstitutional baseline).
As demonstrated in Sec.~\ref{sec:calibration}, maximizing CoFi's accuracy on real-world data can determine an appropriate trust ratio.

    \section{Experiments}
\label{sec:experiments}

We answer the following questions using real-world marine localization from Autonomous Identification System~(AIS) recordings and official charts.
\begin{enumerate}
    \item[\textbf{(Q1)}] How can we encode expert knowledge about marine vessels, like traffic regulations, into a Constitution?
    \item[\textbf{(Q2)}] What are the computational demands of integrating neuro-symbolic reasoning into Bayesian estimation? 
    \item[\textbf{(Q3)}] Can we learn how much to trust marine vessels given the Constitution and historical AIS data?
    \item[\textbf{(Q4)}] Does a trust-calibrated CoFi lead to improved tracking accuracy compared to the unconstitutional baseline?
\end{enumerate}

\subsection{Experimental Setup}
\label{sec:experimental_setup}

Throughout the experiments, we assume a constant velocity model $\mathbf{x}_t = (\mathbf{p}_t, \mathbf{v}_t)$ in a two-dimensional space, i.e., $\mathbf{p}_t, \mathbf{v}_t \in \mathbb{R}^2$ and $\dot{\mathbf{p}}_t = \mathbf{v}_t$, where only the position $\mathbf{p}_t$ is measured.
Hence, given the difference $\delta t$ between time steps, the dynamic system can be written as
\begin{equation*}
    \label{eq:f_and_h}
    \mathbf{x}_t = \left[ \begin{array}{cc}
         1 & \delta t  \\
         0 & 1  
    \end{array} \right] \mathbf{x}_{t-1} \;\; \text{ and } \;\;
    \mathbf{z}_t = \left[ \begin{array}{cc}
         1 & 0
    \end{array} \right] \mathbf{x}_t.
\end{equation*}
All experiments were conducted on an AMD Ryzen Threadripper 1950X 16-Core processor with 128GB of memory.
Where applicable, we report the means and standard deviations obtained from multiple runs with varying seeds.

\subsection{A Constitution for Maritime Traffic Tracking}
\label{sec:modeling}

We address \textbf{(Q1)} in three steps.
First, we integrate messages retrieved via AIS~\cite{series2014technical,vesseldata}, transmitting navigation and ship data in marine applications, into the Constitution's perception subprogram $\mathcal{P}_t$ and trust features $\bm{\psi}$.
Second, common safety and navigational constraints considering the vessel's purpose were encoded into the agents' background knowledge $\mathcal{B}_t$.
Finally, the environment $\mathcal{E}_t$ is represented by a StaR Map processing official Electronic Navigation Charts~(ENC) provided by the United States National Oceanic and Atmospheric Administration~(NOAA).
Figure~\ref{fig:star_maps_parameters} shows the resulting spatial relations, detailing geographic and navigational features surrounding the New York Harbor area.
The composition of $\mathcal{P}_t$, $\mathcal{B}_t$, and $\mathcal{E}_t$ is shown in Listing~\ref{listing:marine_model}.

\begin{figure}
    \centering
    \begin{subfigure}{0.52\linewidth}
        \includegraphics[width=\textwidth]{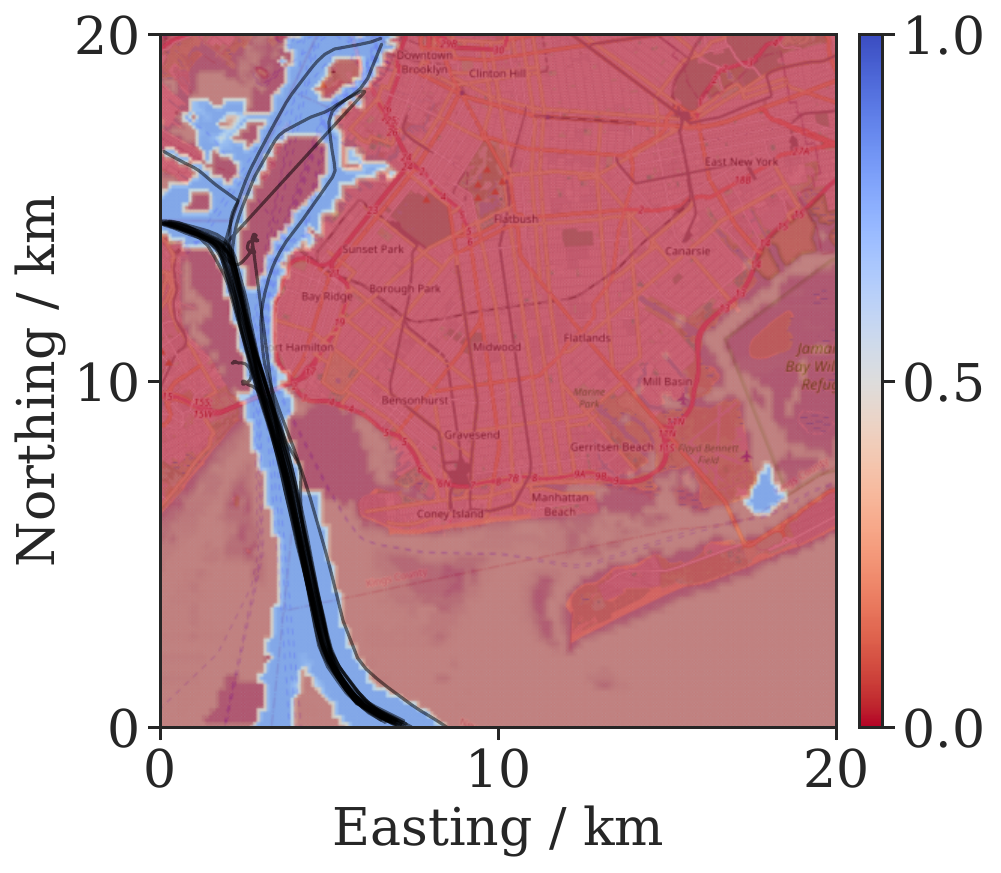}
        \caption{Cargo}
    \end{subfigure} 
    \hfill
    \begin{subfigure}{0.45\linewidth}
        \includegraphics[width=\textwidth]{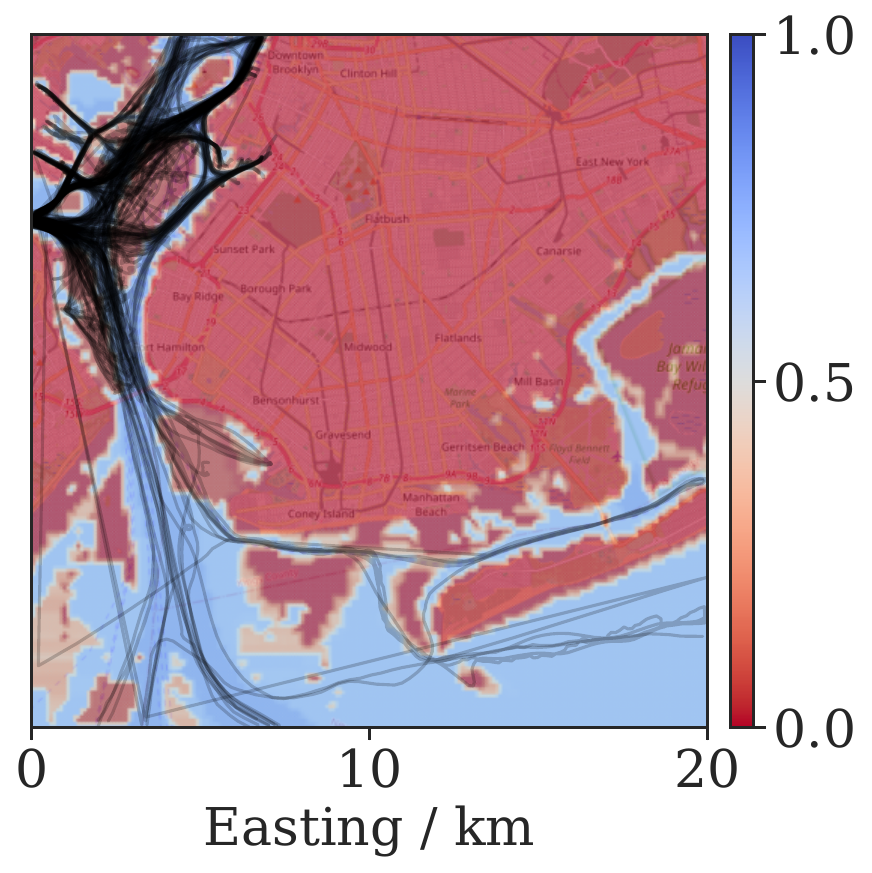}
        \caption{Towing}
        \label{fig:towing_landscape}
    \end{subfigure}
    \caption{
        \textbf{Marine traffic over constitutional probabilities:}
        Here, we show real-world AIS traces from (a) often compliant cargo and (b) often incompliant towing vessels on top of the probabilities for each to satisfy the constitution.
    }
    \label{fig:vessel_types}
\end{figure}

\begin{figure}
    \centering
    \includegraphics[width=\linewidth]{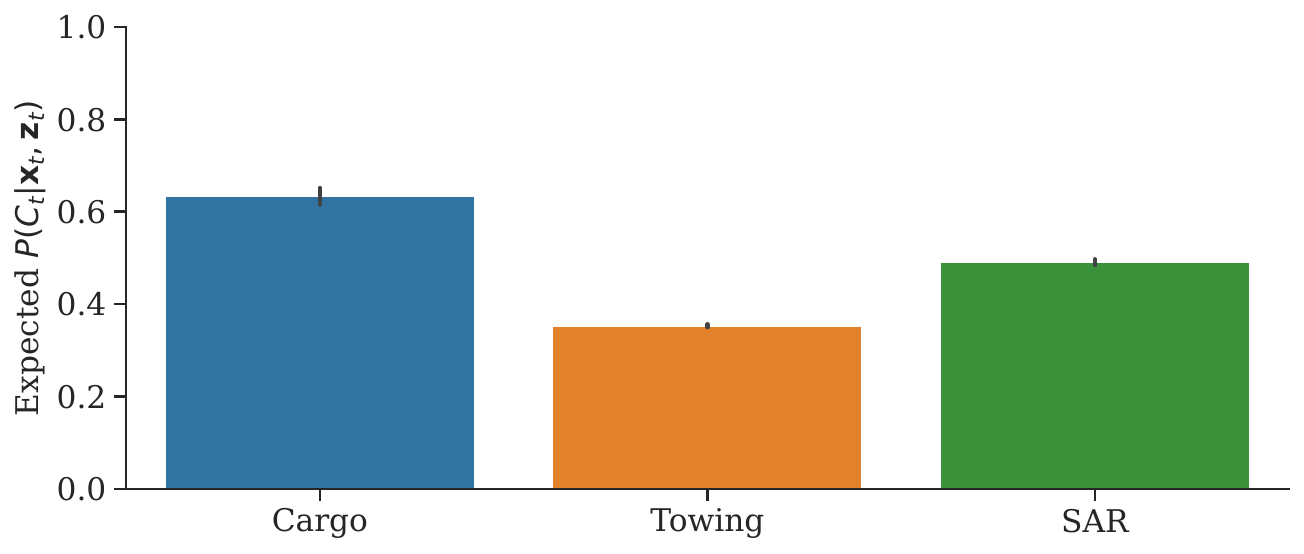}
    \caption{
        \textbf{Trust depending on observations:}
        Not every tracked agent can be expected to satisfy the assumed constitution.
        Here, the average value of $P(C_t | \mathbf{x}_t, \mathbf{z}_t)$ is shown aggregated for different vessel types.
    }
    \label{fig:avg_constitution}
\end{figure}

We solve Listing~\ref{listing:marine_model} for the \textit{constitution(X, Z)} clause across the $20$ $\times$ \SI{20}{\kilo\meter\squared} area by plugging in the respective StaR Maps parameters and observations.
We show how this Constitution successfully models the environment and movement of vessels in the area in Figure~\ref{fig:vessel_types}.
One can see how, although some trajectories lead through assumed incompliant areas, the traffic patterns overall follow the model.
Similarly, Figure~\ref{fig:avg_constitution} shows how the probability of agents sticking to the Constitution depends on, e.g., their employment type.
Note how, for example, towing vehicles are overall less likely to follow the Constitution, as seen in Figure~\ref{fig:towing_landscape}.
We will revisit this observation in Section~\ref{sec:calibration} when determining appropriate trust ratios.

\begin{listing}
    \includegraphics[width=0.9\linewidth]{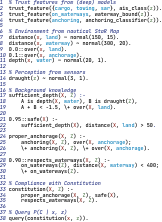}
    \caption{
        A marine vessel's constitution, modeling compliance with safety and traffic rules.
    }
    \label{listing:marine_model}
\end{listing}

\subsection{Resource Requirements}

To address \textbf{(Q2)}, we (i) break down the costs of computing a StaR Map from ENCs and (ii) compare the runtime of tracking vessel motion with a standard Particle Filter and with CoFi.
Computing StaR Maps at a resolution of $100 \times 100$ points for each probabilistic spatial relation took $630.69 \pm 8.51$ seconds.
When querying the Constitution, we employ linear interpolation for parameters not on the grid.
While we compute StaR Maps parameters once in advance, more dynamic applications may need updates due to environmental changes, e.g., a \textit{distance} to a moving target.

The StaR Map parameters have been visualized in Fig.~\ref{fig:star_maps_parameters}, including the \textit{distance} and \textit{over} relations analogously to the original paper and a new \textit{depth} relation introduced here.
While both mean and standard deviation were computed as described in Section~\ref{sec:starmaps}, we visualized the probabilistic spatial relations through their expected values.

We implement CoFi as a Constitutional Particle Filter using sequential Monte Carlo approximation with $2000$ particles.
This allows us to compare CoFi to a common baseline, i.e., we effectively ablate the evaluation of the Constitution and its guidance for the tracking process by controlling for the influence of computing the Constitutional likelihood.

Without the neuro-symbolic reasoning of CoFi, a standard Particle Filter update can be obtained in $0.004 \pm 0.001$ seconds. 
By sampling $100$ times locally according to the particle distribution, CoFi can run constitutional belief updates in $0.829 \pm 0.024$ seconds.
Else, one may precompute the nearby scalar field $P(C_t | \mathbf{x}_t, \mathbf{z}_t)$ if the environment and measurements are sufficiently static.
Then, CoFi's updates run near baseline speed in $0.007 \pm 0.001$ seconds.
Hence, while CoFi can easily keep up in the maritime context, optimizations are necessary for highly dynamic settings.

\subsection{Calibrating the Trust Ratio}

We resolve \textbf{(Q3)} by modeling a marine Constitution as laid out in Section~\ref{sec:modeling}, considering three trust features.
Namely, the type of the vessel extracted from its AIS report, whether it is bound to waterways through a heuristic rule, and a deep learning classifier deciding whether it is currently anchoring.
To this end, we access real-world AIS recordings provided by NOAA detailing, among others, the vessel's position and characteristics, such as high-level usage type or draft.

We can henceforth observe the alignment of individual agents' tracks with the Constitution as previously discussed in Section~\ref{sec:modeling} and visualized by Figures~\ref{fig:vessel_types} and \ref{fig:avg_constitution}.
Note how CoFi decides the trust level $\tau$ based on all three trust features, not only on the shown vessel type.
Here, without loss of generality, we consider the trust features and the appropriate trust to be time-invariant throughout a recorded journey.

\begin{figure}
    \centering
    \includegraphics[width=\linewidth]{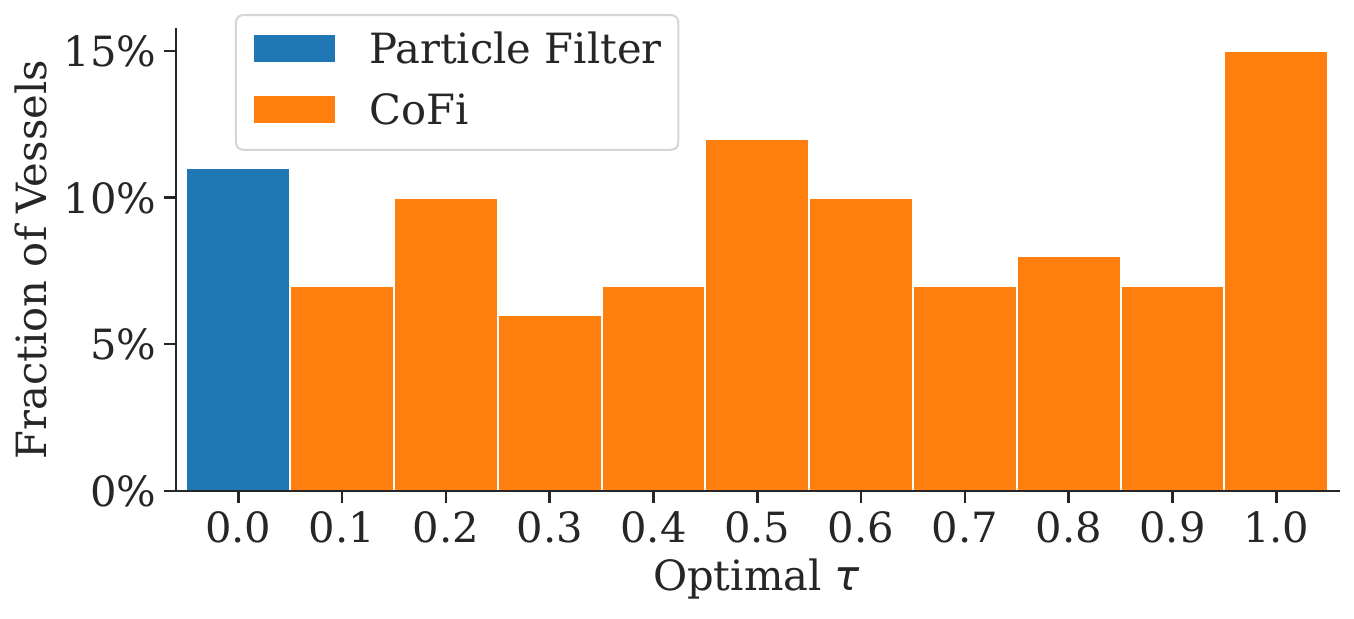}
    \caption{
        \textbf{A majority of vessels can be trusted:}
        Here, the rate at which a specific value of $\tau$ is optimal (given observed trust features) is shown for the entire dataset.
        Note how CoFi rarely needs to fall back to its internal Particle Filter. 
    }
    \label{fig:tau_selection}
\end{figure}

CoFi chooses $\tau$ to maximize its tracking accuracy for agents with the respective trust features.
In our experiments, we perform this computation as an offline learning task on historical AIS data, comparing performance across discrete choices for $\tau$.
In Figure~\ref{fig:tau_selection}, we demonstrate the distribution of optimal choices for $\tau$ across the entire dataset population.
It becomes evident again that, although simple, the assumed Constitution matches most agents' behavior.
That is, about $89\%$ of agents are tracked best for a trust $\tau > 0$, benefiting from CoFi's improved estimation, as further discussed next.

\subsection{Constitutional Filtering}
\label{sec:calibration}

\begin{figure}
    \centering
    \includegraphics[width=\linewidth]{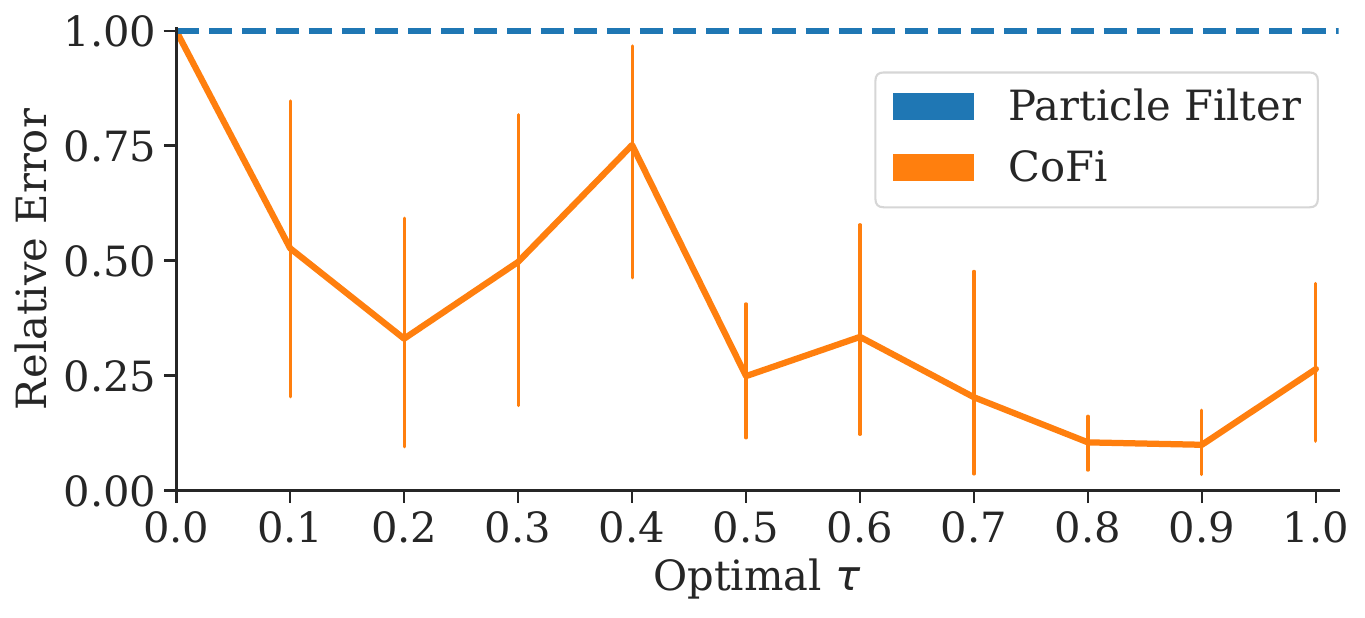}
    \caption{
        \textbf{CoFi adapting to appropriate trust:}
        We show the relative mean absolute error, comparing CoFi's performance with the baseline Particle Filter.
        Once a vessel profits from $\tau > 0$, CoFi achieves greatly improved accuracy.
    }
    \label{fig:accuracy}
\end{figure}

Finally, to answer \textbf{(Q4)}, we employ the obtained optimal trust $\tau$ for respective trust features $\bm{\psi}$ to assess how CoFi improves estimation accuracy if properly calibrated.
Depending on the learned optimal trust for a vessel's trust features, the Constitution of CoFi improves accuracy or recovers baseline performance.
So, if trust features indicate the vessel does not comply with the assumed model and the optimal $\tau$ is $0$, the belief update equals a Particle Filter without Constitution.
As Figure~\ref{fig:accuracy} shows, CoFi provides more accurate tracking as soon as $\tau > 0$, leveraging the information provided by the Constitutional likelihood in CoFi's belief update.

Note that there are limitations to these results, considering the choice of the Constitution.
While a trust-calibrated CoFi clearly can lead to improved results and assures an upper bound to its error, not every environment and application will allow for formulating a Constitution that clearly separates likely and unlikely states.
For example, while roads, pavements, and traffic lights give way to predicting traffic participants' motion in an urban environment, the same cannot be said for areas where free motion is predominant. 

    \section{Discussion}
\label{sec:discussion}

We propose the Constitutional Filter~(CoFi), a neuro-symbolic Bayesian estimation scheme based on a model of an agent's internal rules over their behavior.
CoFi improves traditional filtering approaches by reasoning on an agent's Constitution.
In experiments on real-world map and localization data, we have demonstrated how the Constitution can substantially improve the filter's accuracy and how its impact can be tuned to an agent's individual behavior.
Furthermore, through the symbolic nature of the Constitution written in probabilistic first-order logic, an interpretable and adaptable interface to the filtering process is provided.

Although CoFi brings several advantages, as shown in Section~\ref{sec:experiments}, they depend on application-specific background knowledge which must be obtained first.
While we have shown a hand-crafted Constitution, real-world applications may need to explore options such as rule-learning or automated code generation via Large Language Models.
Furthermore, prediction with CoFi entails additional computational costs, possibly prohibiting its application in environments where particularly rapid inference is key.

In the future, end-to-end learning with the agent's Constitution, i.e., learning the rules' conditional probabilities and bodies from data, as well as the weights of neural models feeding into the Constitution, could be explored.
Similarly, the automatic discovery of an agent's reasons for non-compliance with the assumed Constitution is a promising direction forward, i.e., learning if the Constitution itself is faulty or if the agent deliberately does not comply.

    \section*{Acknowledgments}
    Simon Kohaut gratefully acknowledges the financial support from Honda Research Institute Europe.
    This work received funding from the German BMBF project~\enquote{KompAKI} within the \enquote{The Future of Value Creation -- Research on Production, Services and Work} program (grant 02L19C150), managed by the PTKA.
    The TU Eindhoven authors received support from their Department of Mathematics and Computer Science and the Eindhoven Artificial Intelligence Systems Institute.
    Map data \copyright~OpenStreetMap contributors, licensed under the ODbL and available from \href{https://www.openstreetmap.org}{openstreetmap.org}.
    
    \bibliographystyle{IEEEtran}
    \bibliography{IEEEabrv, references.bib}
\end{document}